\documentclass[nolinenumbers]{trbunofficial}
\usepackage{wrapfig}
\usepackage{booktabs}
\usepackage[hidelinks]{hyperref}
\usepackage{graphicx}
\usepackage{float}
\usepackage{multirow}
\usepackage{multicol}
\usepackage{wrapfig}
\usepackage{graphicx}
\usepackage[hidelinks]{hyperref}
\usepackage[percent]{overpic}
\usepackage{graphicx}
\usepackage{tikz}   

\makeatletter
\renewcommand{\fnum@figure}{\bfseries Figure~\thefigure}
\renewcommand{\fnum@table}{\bfseries TABLE~\thetable}
\makeatother

\usepackage[
  authordate,
  backend=biber,
  natbib=true,
  giveninits=false,
  maxbibnames=99,
  minbibnames=99,
  doi=true,
  url=true,
  isbn=false
]{biblatex-chicago}

\captionstyle{\raggedright}
\normalcaptionwidth

\begin{document}

\title{Schedule-Informed Temporal Fusion Forecasting of Hourly Airport Security-Checkpoint Throughput}

\TRBauthor{Yinxiao Zhang}{School of Aviation and Transportation Technology, Purdue University}{zhan3110@purdue.edu}[West Lafayette, Indiana, 47907][https://orcid.org/0009-0007-6653-5891]
\TRBauthor{Sen Wang, Ph.D.*}{Department of Geography, The Ohio State University}{wang.18872@osu.edu}[Columbus, Ohio, 43210][https://orcid.org/0000-0001-9891-0325]
\TRBauthor{Yi Gao, Ph.D.}{College of Aeronautics and Engineering, Kent State University}{ygaoaero@kent.edu}[Kent, Ohio, 44240][https://orcid.org/0000-0002-5996-3288]

\AuthorHeaders{Zhang, Wang, and Gao}

\maketitle









\newpage
\section{Abstract}

\noindent\textbf{Objectives:}
Checkpoint staffing decisions depend not only on how many passengers are screened but on when screening load materializes. Yet flight schedules timestamp departures, whereas passengers reach the checkpoint earlier. We develop and evaluate a framework that transforms known flight schedules into temporally aligned signals for forecasting hourly checkpoint throughput.\\

\noindent\textbf{Methods:}
Using 2023--2024 hourly Transportation Security Administration throughput and Cirium Diio flight schedules for Hartsfield--Jackson Atlanta International Airport, we trained and validated models chronologically and held out July--December 2024 for testing. Domestic and international seat capacity was distributed across pre-departure hours with truncated Poisson lead-time kernels. A Temporal Fusion Transformer fused these arrival-intensity signals with historical throughput, scheduled activity, and calendar and seasonal variables. Direct six-hour and recursive 24--96 hour forecasts were benchmarked against recurrent neural network and long short-term memory models with identical inputs across five random seeds.\\

\noindent\textbf{Findings:}
For direct six-hour forecasts, the proposed model achieved a weighted mean absolute percentage error of 9.33\%, compared with 12.16\% for the recurrent neural network and 11.37\% for long short-term memory, and yielded the lowest errors above the 75th and 90th throughput percentiles. With six-hour recursive updates, this metric remained between 10.60\% and 11.04\% across 24--96 hour horizons, without monotonic aggregate error growth; however, longer horizons had fewer valid forecast origins.\\

\noindent\textbf{Novelty:}
Rather than treating scheduled departures as contemporaneous demand, the framework converts them into temporally distributed screening-load signals available at forecast issuance. This schedule-to-checkpoint representation encodes pre-departure lead-time structure in interpretable multi-horizon forecasting without individual passenger--flight matches.\\

\noindent\textbf{Practical Applications:}
The forecasts can inform advance staffing, lane opening decisions, and multiday checkpoint planning. They should be combined with local capacity, staffing, queue, and wait time information because observed throughput reflects realized processing rather than unconstrained passenger arrivals.

\newpage

\section{Introduction}
\label{sec:introduction}

Airport security screening is a critical bottleneck in airport terminal operations because it governs passenger access to the secured side of the terminal and directly affects passenger processing, queue formation, staffing requirements, and service reliability. In the United States, Transportation Security Administration (TSA) Domestic Aviation Operations is responsible for nearly 440 federalized airports and screens more than two million passengers daily \citep{PerformanceGov2026}. At this scale, even short periods of demand–capacity imbalance can produce long queues, underutilized or overextended screening lanes, and degraded passenger experience. Because screening capacity is adjusted through discrete operational decisions, such as opening lanes and assigning Transportation Security Officers, accurate demand prediction is essential for managing checkpoint performance \citep{hanumantha2020demand}.

The usefulness of passenger-volume forecasting depends not only on predictive accuracy, but also on whether the forecast is aligned with the temporal scale of checkpoint decision-making. Daily passenger totals may describe the overall level of airport activity, but they provide limited guidance for managing within-hour/day screening pressure. At security checkpoints, congestion develops from the interaction between time-varying passenger inflows and available screening capacity, a relationship consistent with the queuing theory commonly used to analyze service systems and airport operations \citep{hu2023passenger,jacquillat2015integrated}. Staffing levels, lane-opening decisions, and queue-management actions must therefore anticipate demand fluctuation before queues accumulate. This creates a forecasting problem that is both temporal and operational: passenger demand must be estimated at a sufficiently fine resolution to support intraday checkpoint management and proactive resource allocation. 

One of the central challenges in TSA checkpoint forecasting is that flight schedules are informative about future screening demand but are not temporally aligned with the moment at which passengers enter the screening process. Scheduled departures indicate when passenger trips leave the airport system, whereas checkpoint demand is generated earlier as passengers arrive, complete pre-security activities, and proceed through screening before boarding. The lead time between checkpoint arrival and flight departure can vary across flight types, traveler characteristics, airport layout, airline guidance, and perceived uncertainty about processing times. As a result, a scheduled departure contributes to a distributed pattern of screening demand over preceding time intervals rather than only during its departure hour. This temporal displacement motivates schedule-derived representations that translate future flight activity from departure-time information into expected pre-departure screening-load profiles.

Existing research provides important foundations for this problem. Passenger-demand forecasting studies have explored statistical, simulation-based, machine-learning, and deep-learning approaches for estimating air travel volumes across different temporal scales \citep{alsayed2026air,ruiz2020simulation,ghandeharioun2023exploring,kanavos2021deep}. Related studies on airport terminal and checkpoint operations have also shown that flight schedules, passenger-arrival profiles, and historical passenger-flow patterns can support terminal-flow prediction, checkpoint-demand estimation, and workforce allocation decisions \citep{lin2023prediction,viana2024explainable,alodhaibi2019impact}. Building on this literature, this study further integrates schedule-based arrival logic with multi-horizon forecasting by jointly using historical throughput patterns, known future schedule information, and temporal covariates. This integration is important because checkpoint forecasts must capture not only the expected magnitude of screening demand, but also the hours in which that passenger arrival pattern is likely to materialize.

To operationalize this integration, we develop a schedule-informed framework for hourly TSA checkpoint throughput prediction. Leveraging the available data, we use observed hourly TSA throughput as the measure of realized screening load, while the unobserved pre-screening arrival process is represented through flight-schedule-derived covariates. Scheduled departures are transformed into pre-departure arrival-intensity features using truncated Poisson lead-time kernels. These features are then combined with historical TSA throughput, flight schedule measures, calendar variables, and seasonal time features within a Temporal Fusion Transformer framework, which supports interpretable multi-horizon forecasting with historical observations, known future inputs, and time-varying covariates \citep{lim2021temporal}. This formulation is operationally essential because the objective is not only to estimate the volume of screening demand, but also to assign that demand to the correct hours. Even when total passenger volume over a planning window is forecast reasonably well, misplacing demand across hours can compromise the usefulness of the forecast for checkpoint planning.

This study makes three contributions to the airport operations. First, it frames hourly TSA throughput forecasting as a checkpoint screening-load management problem, rather than a general passenger-volume forecasting task. Second, it develops a schedule-derived arrival-intensity representation that addresses the temporal mismatch between scheduled departures and checkpoint demand. Third, it integrates these features into a multi-horizon attention-based transformed framework and evaluates performance under direct, rolling, longer-horizon, and high-throughput forecasting settings. Together, these contributions advance a forecasting approach that connects known future flight schedules, historical throughput patterns, and operationally meaningful checkpoint-demand prediction.

The remainder of this study is organized as follows. Section 2 reviews and synthesizes the relevant literature. Section 3 introduces the research scope, analytical framework, and data used to construct the variables. Section 4 presents the empirical findings. Section 5 discusses the implications of the findings, acknowledges the study’s limitations, and identifies directions for future research. Section 6 provides concluding remarks and summarizes the study’s key contributions.

\section{Literature Review}

\subsection{Passenger-flow forecasting}

Airport passenger forecasting has traditionally been used to support planning decisions at airport, route, regional, or national scales. Much of this literature estimates passenger volumes using econometric, statistical, time-series, or machine-learning approaches \citep{tsui2014forecasting,kanavos2021deep,alsayed2026air}. These studies provide important methodological foundations for modeling air travel demand, but their forecasting targets are often broader than the operational problem addressed in this study. For checkpoint management, the central concern is not only how many passengers will travel, but when passengers will enter the security-screening process.

Recent airport passenger-flow studies have moved closer to this operational perspective. \citet{hopfe2024short} compare conventional time-series models with neural-network approaches for short-term airport security-checkpoint passenger-flow forecasting at major U.S. airports, showing that recurrent neural networks can improve forecasting performance, particularly when exogenous variables are included. \citet{guo2022forecasting} develop a machine-learning framework for real-time distributional forecasting of airport transfer passenger flows, including arrivals at immigration and security areas. \citet{viana2024explainable} propose an explainable model for predicting incoming security-checkpoint passengers at Cincinnati/Northern Kentucky International Airport, emphasizing the relevance of short-term forecasts for TSA officer scheduling and resource planning.

A related stream of research connects passenger-flow prediction directly to checkpoint resource allocation. \citet{hanumantha2020demand} combine demand prediction with dynamic workforce allocation to improve airport screening operations, while \citet{brun2025schedule} optimize security-checkpoint opening schedules and staff allocation using predicted passenger flows. These studies demonstrate that checkpoint forecasts have practical value when they support decisions about staffing, lane openings, and service reliability. However, they also imply a more specific forecasting challenge: to support these decisions, predicted demand must be assigned to the correct operating periods. This raises the question of how future flight schedules should be translated into the timing of checkpoint arrivals.

\subsection{From flight schedules to checkpoint-arrival intensity}

Flight schedules are a key source of known future information for airport passenger-flow forecasting. Scheduled departures provide advance information about outbound activity, including departure times, flight frequency, aircraft size, and the temporal concentration of passenger-generating events. However, departure time is not equivalent to checkpoint-arrival time. Departing passengers typically arrive before their flights, complete pre-security activities such as check-in or baggage drop, and enter screening over a pre-departure interval. The timing of this process varies by passenger type, airline, travel purpose, baggage status, airport layout, and perceived processing uncertainty.

Research on passenger-arrival patterns provides the basis for modeling this temporal displacement. \citet{alodhaibi2019impact} show that arrival patterns affect outbound airport processes, including check-in, security screening, and immigration. \citet{postorino2019airport} model airport passenger arrivals using earliness-arrival functions, emphasizing that arrivals should be represented relative to scheduled departure times. \citet{lin2023prediction} show that flight arrangements can forecast passenger distributions across terminal areas, but passenger peaks occur at different times in check-in, security, and departure spaces. These studies suggest that flight schedules are informative only after they are translated into the timing of passenger movement through the terminal. The show-up-profile literature makes this schedule-to-arrival link explicit. A show-up profile describes the probability distribution of passenger arrivals before a scheduled service event. \citet{bernstein2026frontiers} show how such profiles can be combined with known flight schedules to generate aggregate passenger-arrival forecasts. This idea is particularly relevant for airport security screening because detailed passenger-flight matching data or individual-level arrival records are often unavailable. In such settings, arrival behavior must be inferred from observable operational data and represented through schedule-based arrival profiles.

This motivates the schedule-derived arrival-intensity features used in this study. Observed TSA throughput measures realized screening load, but it does not identify the flight-specific arrival behavior of individual passengers. Simple hourly departure counts may capture broad activity levels, but they can misplace demand because passengers are screened before departure. A more appropriate representation distributes the influence of each scheduled departure over preceding time intervals. Accordingly, this study transforms scheduled departures into pre-departure arrival-intensity features using truncated Poisson lead-time kernels.

\subsection{Multi-horizon forecasting}

TSA throughput forecasting is a multi-horizon time-series problem with heterogeneous inputs. Historical throughput captures recurring screening-load patterns, including time-of-day peaks, weekday variation, seasonality, and airport-specific operating rhythms. Flight schedules provide known future information, while schedule-derived arrival-intensity features translate those schedules into expected screening-load timing. Calendar and seasonal time features provide additional temporal structure. A suitable model must therefore combine historical observations, known future covariates, and time-varying predictors across multiple horizons.

Deep-learning methods are increasingly used in transportation forecasting because they can represent nonlinear temporal relationships among multiple predictors \citet{polson2017deep, choi2021artificial,nguyen2018deep}. In airport passenger-flow forecasting, neural-network models have shown promise for short-term prediction, especially when exogenous variables are included \citep{hopfe2024short,monmousseau2020predicting,guo2022forecasting}. However, checkpoint forecasting requires more than flexible nonlinear prediction. Because forecasts support staffing and lane-opening decisions, the model should also accommodate known future operational inputs and provide interpretable information about the factors driving predicted screening load. 

The Temporal Fusion Transformer is well suited to this requirement. \citet{lim2021temporal} introduced the Temporal Fusion Transformer as an interpretable architecture for multi-horizon time-series forecasting. The model can incorporate historical time-varying variables, known future inputs, and static or slowly varying covariates, while its variable-selection and attention mechanisms help identify important predictors and temporal dependencies. These characteristics match the checkpoint-forecasting problem: historical throughput represents realized screening dynamics, future flight schedules represent passenger-generating events, and arrival-intensity features translate those events into expected screening-load timing.

\subsection{Summary of the existing literature}

Overall, prior studies provide foundations in airport passenger-flow forecasting, passenger-arrival modeling, and multi-horizon prediction, these streams are rarely integrated for TSA throughput forecasting. Existing studies often demonstrate the value of machine-learning forecasts or schedule-based arrival profiles, but less attention has been given to combining schedule-derived arrival intensity with interpretable multi-horizon forecasting. This study addresses that gap by developing a schedule-informed Temporal Fusion Transformer framework that integrates historical TSA throughput, flight-schedule measures, calendar variables, seasonal time features, and truncated Poisson arrival-intensity features. The contribution is both methodological and operational: it reframes TSA throughput forecasting as a screening-load management problem in which correctly timing demand is as important as estimating total volume.

\section{Methodology}
\label{sec:methodology}

\subsection{Research Setting}
\begin{figure}[H]
    \centering
    \includegraphics[width=1\linewidth]{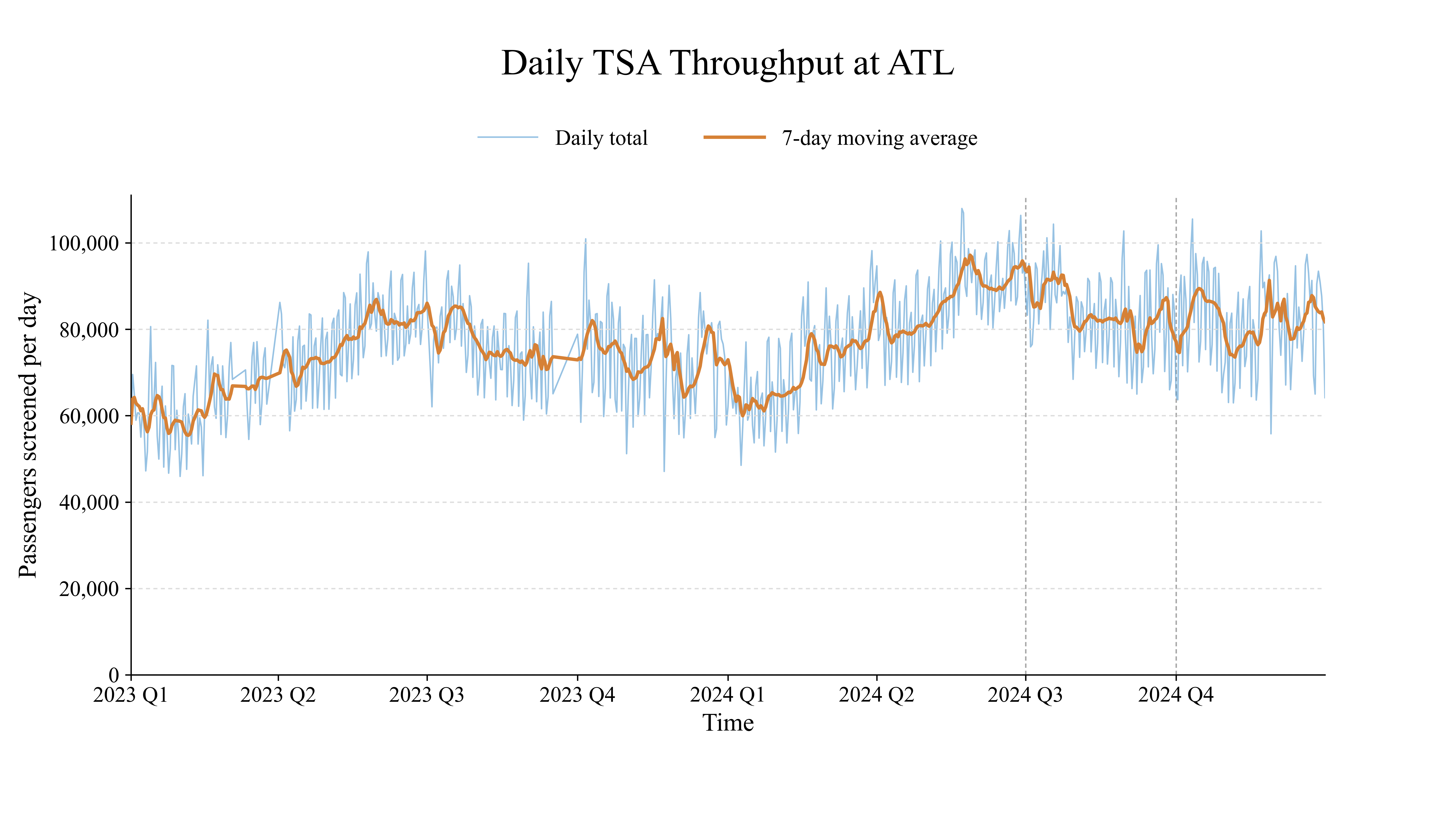}
    \caption{Daily TSA Throughput at ATL}
    \label{fig:ATL_count}
\end{figure}

The empirical analysis focuses on Hartsfield--Jackson Atlanta International Airport (ATL). ATL was selected because its high passenger volume, extensive flight schedule, and continuous checkpoint activity provide an appropriate setting for evaluating hourly checkpoint-throughput forecasts across short- and long-term planning horizons (shown in Figure~\ref{fig:ATL_count}). The dataset covers the 2023 and 2024 calendar years and combines hourly TSA checkpoint-throughput data released by the Transportation Security Administration \citep{tsa_throughput_data} with flight-schedule information obtained from Cirium Diio \citep{cirium_diio}. Flight-schedule records are aggregated to the hourly level to construct measures of scheduled departures, scheduled seat capacity, and schedule-derived passenger-arrival intensity.

Let \(Y_{a,t}\) denote the observed number of passengers screened by TSA at airport \(a\) during hour \(t\). This variable is the prediction target and is interpreted as realized checkpoint screening load. It does not directly measure latent passenger arrivals before screening, checkpoint capacity, queue length, or passenger wait time. Although the forecasting framework is expressed generally for airport \(a\), the empirical application in this study focuses on ATL.

At forecast origin \(t\), the model uses an encoder window containing the previous \(L\) hours of observed throughput and historical covariates to predict throughput over the subsequent \(H\) hours. The model also receives covariates that are known over the future prediction window. Let \(\mathbf{X}_{a,s}\) denote the covariate vector for airport \(a\) at hour \(s\). Accordingly, \(\mathbf{X}^{\mathrm{obs}}_{a,t-L+1:t}\) and \(\mathbf{X}^{\mathrm{known}}_{a,t+1:t+H}\) denote the sequences of historically observed and future-known covariates over the encoder and prediction windows, respectively. The forecasting problem is represented as:

\begin{equation}
\hat{\mathbf{Y}}_{a,t+1:t+H}
=
f_{\theta}
\left(
\mathbf{Y}_{a,t-L+1:t},
\mathbf{X}^{\mathrm{obs}}_{a,t-L+1:t},
\mathbf{X}^{\mathrm{known}}_{a,t+1:t+H}
\right),
\label{eq:forecasting_framework}
\end{equation}

where \(\hat{\mathbf{Y}}_{a,t+1:t+H}\) is the vector of predicted hourly throughput, \(L\) is the encoder-window length, \(H\) is the prediction horizon, and \(f_{\theta}\) denotes the fitted forecasting model. 

Historical observed inputs include lagged TSA throughput and target-derived temporal features. Future-known inputs include calendar and seasonal variables, scheduled flight activity, scheduled seat capacity, and the schedule-derived arrival-intensity measures described in the following subsection. Features are assigned to the observed or future-known input sets according to their availability at the forecast origin. Only schedule information available before the corresponding forecast origin is used to construct future covariates.

The dataset was divided chronologically into contiguous training, validation, and held-out evaluation periods to preserve temporal ordering and prevent look-ahead bias. Observations from January 1, 2023, through March 31, 2024, were used for model estimation. Data from April 1 through June 30, 2024, were used for hyperparameter selection, early stopping, checkpoint selection, and rolling-configuration selection. The remaining observations, from July 1 through December 31, 2024, were reserved exclusively for final out-of-sample evaluation. All preprocessing parameters were estimated using the training data and then applied without re-estimation to the validation and evaluation periods. The evaluation period was not used to select model hyperparameters, forecasting configurations, or checkpoints.

Forecasting samples were assigned to a partition according to their forecast origins. For forecast origins near the beginning of the validation or evaluation period, the encoder was permitted to use observations from the immediately preceding period because these historical observations would have been available when the forecast was issued. No target observations occurring after a forecast origin were used as model inputs, and samples without complete encoder and prediction windows were excluded.

\subsection{Schedule-Derived Arrival-Intensity Features}

\begin{figure}[h]
    \centering
    \includegraphics[width=1\linewidth]{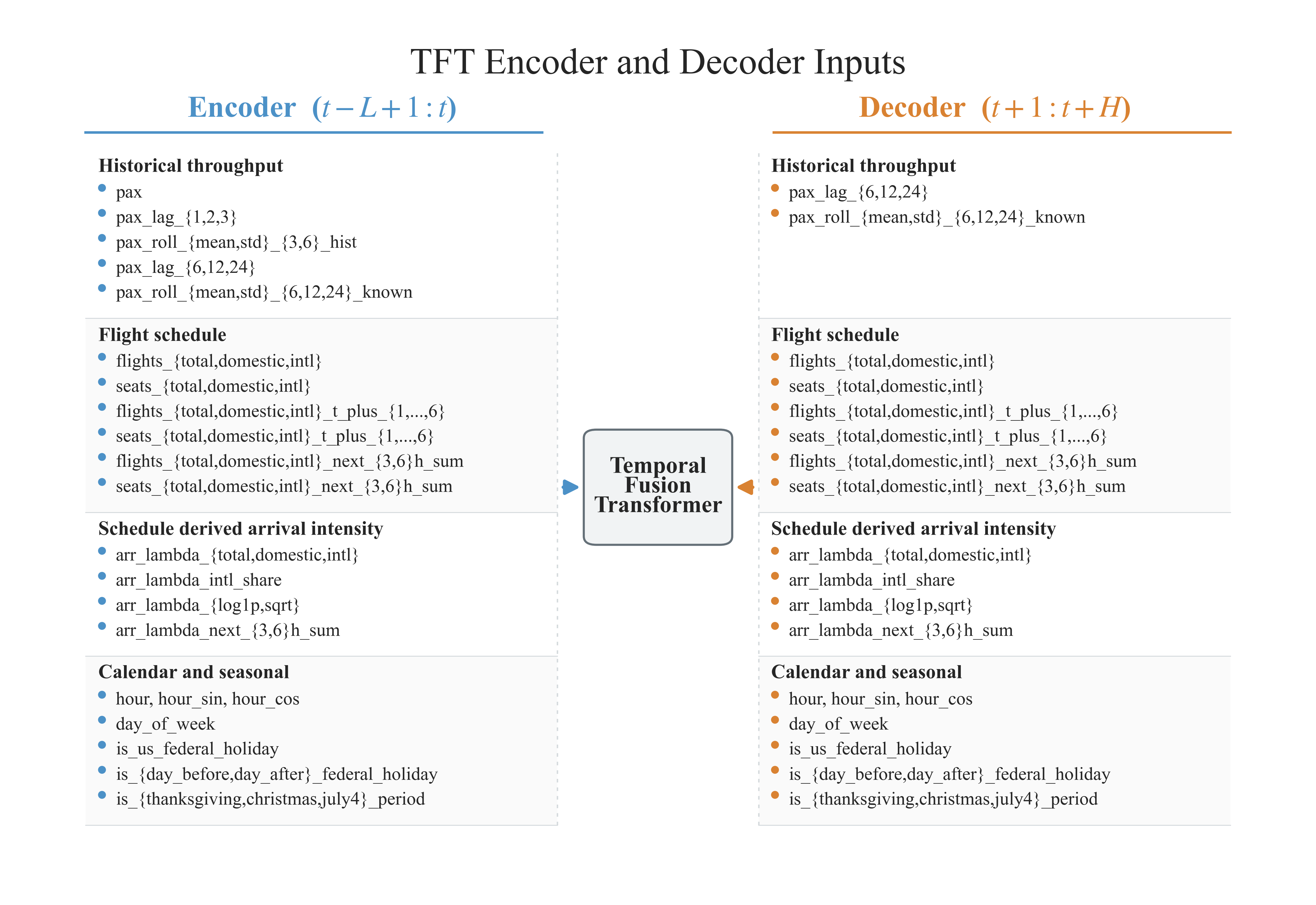}
    \caption{TFT Encoder and Decoder Inputs}
    \label{fig:encoder_decoder}
\end{figure}

Flight schedules provide advance information about future passenger activity, but scheduled departure time is not temporally aligned with the time at which passengers enter the security checkpoint. Passengers generally complete screening before departure, and the associated lead time may differ between domestic and international travel. Directly assigning scheduled flight capacity to the departure hour may therefore place passenger demand later than it is likely to materialize at the checkpoint. To address this temporal mismatch, scheduled flight activity is transformed into pre-departure arrival-intensity features shown in figure Figure~\ref{fig:encoder_decoder}.

Figure~\ref{fig:encoder_decoder} groups the model inputs into historical throughput, flight schedules, schedule-derived arrival intensity, and calendar and seasonal variables. Historical lags and rolling statistics capture short-term persistence and recurring intraday patterns. Flight counts and scheduled seat capacity represent future airport activity and potential passenger volume, while the domestic and international components account for differences in pre-departure arrival patterns. Schedule-derived arrival-intensity features align scheduled departures with earlier checkpoint screening periods, and calendar variables capture recurring hourly, weekly, and holiday effects. Only information available at the forecast origin is supplied to the decoder, thereby preventing post-origin throughput from entering the model.

The construction is motivated by queueing models in which arrivals over fixed time intervals are represented through time-varying arrival intensities \citep{gross2018fundamentals,kleinrock1975queueing}. The Poisson formulation is used only as a lead-time weighting function for feature construction; it does not impose a Poisson distribution on observed TSA throughput.

Let \(j\) index scheduled departing flights, \(a_j\) denote the departure airport, \(g_j \in \{\mathrm{dom},\mathrm{intl}\}\) denote the domestic or international flight category, \(c_j\) denote scheduled seat capacity, and \(\tau_j\) denote the scheduled departure hour. For flight category \(g\), the normalized lead-time weight assigned to hour \(r\) before departure is

\begin{equation}
w_{g,r}
=
\frac{
\mu_g^{r}e^{-\mu_g}/r!
}{
\displaystyle
\sum_{q\in\mathcal{R}}
\mu_g^{q}e^{-\mu_g}/q!
},
\qquad
r\in\mathcal{R},
\label{eq:truncated_poisson}
\end{equation}

where \(\mathcal{R}=\{1,\ldots,R\}\) is the allowable lead-time support and \(\mu_g\) controls the location of the lead-time profile for flight category \(g\). The denominator normalizes the probability mass over the truncated support so that

\[
\sum_{r\in\mathcal{R}}w_{g,r}=1.
\]

The schedule-derived arrival intensity associated with category \(g\) during airport hour \(t\) is then

\begin{equation}
\lambda^{\mathrm{sch},g}_{a,t}
=
\sum_{j:a_j=a,\,g_j=g}
c_j
w_{g,\tau_j-t}
\mathbb{I}
\left(
\tau_j-t\in\mathcal{R}
\right),
\label{eq:schedule_arrival_intensity_group}
\end{equation}

where \(\mathbb{I}(\cdot)\) is the indicator function. Equation \ref{eq:schedule_arrival_intensity_group} distributes the scheduled capacity of each flight across the preceding checkpoint-arrival hours according to its category-specific lead-time profile. The total schedule-derived arrival intensity is

\begin{equation}
\lambda^{\mathrm{sch}}_{a,t}
=
\lambda^{\mathrm{sch},\mathrm{dom}}_{a,t}
+
\lambda^{\mathrm{sch},\mathrm{intl}}_{a,t}.
\label{eq:schedule_arrival_intensity}
\end{equation}

The baseline specification uses \begin{equation}
\left(
\mu_{\mathrm{dom}},
\mu_{\mathrm{intl}}
\right)
=
(1.5,2.5),
\label{eq:baseline_kernel}
\end{equation}

which reflects the expectation that international passengers generally reach the checkpoint earlier than domestic passengers. This configuration is treated as a parsimonious baseline consistent with common airport and airline arrival-time guidance \citep{atl_passenger_security,aa_checkin_arrival}; it is not presented as an empirically estimated passenger show-up distribution. The parameters are specified before evaluation and are not selected by minimizing error on the held-out evaluation period. Alternative domestic and international parameter combinations are examined as sensitivity checks.

\subsection{Forecasting Models}

\subsubsection{TFT Model}
The Temporal Fusion Transformer (TFT) is used as the primary forecasting model because it can jointly represent historical observations, time-varying covariates, known future inputs, and multi-horizon outputs \citep{lim2021temporal}. At each airport hour, the candidate input variables are grouped as:

\begin{equation}
\mathbf{x}_{a,t}
=
\left[
\mathbf{x}^{\mathrm{hist}}_{a,t},
\mathbf{x}^{\mathrm{cal}}_{a,t},
\mathbf{x}^{\mathrm{sch}}_{a,t},
\mathbf{x}^{\mathrm{arr}}_{a,t}
\right],
\label{eq:tft_inputs}
\end{equation}

where $\mathbf{x}^{\mathrm{hist}}_{a,t}$ contains lagged TSA throughput and target-derived temporal features; $\mathbf{x}^{\mathrm{cal}}_{a,t}$ contains calendar and seasonal variables; $\mathbf{x}^{\mathrm{sch}}_{a,t}$ contains scheduled flight activity and seat-capacity measures; and $\mathbf{x}^{\mathrm{arr}}_{a,t}$ contains the domestic, international, and total schedule-derived arrival-intensity features.

Historical TSA throughput and other observed covariates are supplied only through the forecast origin. Calendar variables, scheduled flight activity, scheduled seat capacity, and schedule-derived arrival intensity are treated as future-known inputs because they can be constructed for the prediction window using information available when the forecast is issued. No realized throughput observations occurring after the forecast origin are used as model inputs. As the empirical application focuses on a single airport, no airport-level static covariates are included. 

\begin{figure}[t]
    \centering
    \includegraphics[width=1\linewidth]{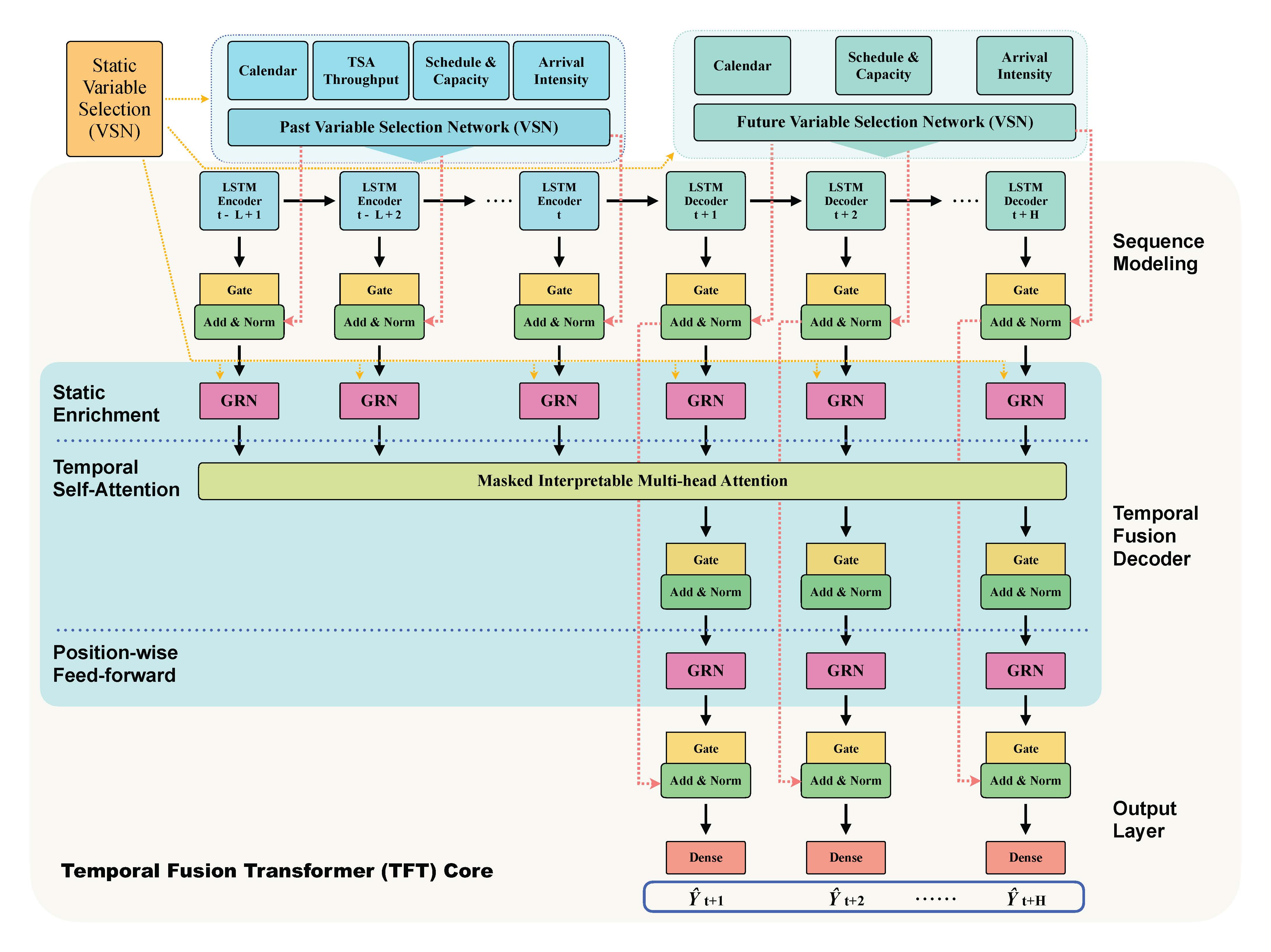}
    \caption{TFT framework}
    \label{fig:TFT_TSA}
\end{figure}

Within the TFT architecture, variable-selection networks assign time-varying importance to the candidate predictors. Recurrent encoder--decoder components represent local sequential relationships, while masked temporal attention identifies historical periods relevant to each forecast horizon. The decoder combines the encoded historical state with future-known covariates and produces a separate hourly throughput prediction for each step from $t+1$ through $t+H$. Although TFT can be used for probabilistic forecasting, the present implementation produces point forecasts and is trained using mean absolute error loss. Figure \ref{fig:TFT_TSA} summarizes the flow of historical throughput, calendar variables, flight schedules, and schedule-derived arrival-intensity features through the forecasting framework.

\subsubsection{RNN and LSTM Baselines}
To provide conventional neural sequence-modeling benchmarks, a recurrent neural network (RNN) and a long short-term memory (LSTM) network are evaluated alongside the TFT model. Both recurrent baselines use the same predictor groups defined in Equation~(6). The models also use the same encoder length, prediction horizon, chronological data partitions, and training objective as TFT, thereby allowing the comparison to focus primarily on differences in model architecture.

The RNN provides a basic recurrent benchmark for modeling temporal dependence in hourly checkpoint throughput. It processes the input sequence in chronological order and updates an internal hidden state at each time step. This hidden state summarizes information from the current input and the preceding sequence and is subsequently used to generate forecasts over the prediction horizon. Through this recurrent structure, the model can represent short-term relationships among recent throughput, temporal patterns, and flight-schedule-related covariates. However, a conventional RNN may have difficulty preserving information across long sequences because the influence of earlier observations can diminish as the hidden state is repeatedly
updated. The model also does not explicitly distinguish the relative importance of individual predictors or historical time steps.

The LSTM is included as a stronger recurrent benchmark designed to address these limitations. It extends the conventional RNN through an internal memory state and a set of input, forget, and output gates. These gates regulate which information is added to memory, retained from previous time steps, and transferred to the hidden state. As a result, the LSTM can preserve relevant temporal information over longer encoder windows and is better suited to capturing recurring daily and weekly throughput patterns. In this study, the LSTM follows the same forecasting structure and receives the same historical and future-known inputs as the RNN. Nevertheless, neither recurrent baseline includes the explicit variable-selection networks, static enrichment, or masked temporal-attention mechanisms used by TFT. Their inclusion therefore provides a controlled assessment of whether basic recurrent processing or gated memory alone can account for the forecasting performance of the broader temporal-fusion architecture.

\subsection{Experimental Design and Forecast Evaluation}

Forecast accuracy is evaluated across all airport hours and forecast origins in the held-out evaluation period. Let \(i=1,\ldots,N\) index the evaluated airport-hour predictions. The primary performance measures are mean absolute error (MAE) and weighted mean absolute percentage error (WMAPE), while root mean squared error (RMSE) is used as a secondary measure:

\begin{align}
\mathrm{MAE}
&=
\frac{1}{N}
\sum_{i=1}^{N}
\left|Y_i-\hat{Y}_i\right|,
\\
\mathrm{WMAPE}
&=
\frac{
\sum_{i=1}^{N}
\left|Y_i-\hat{Y}_i\right|
}{
\sum_{i=1}^{N}Y_i
}
\times 100\%,
\\
\mathrm{RMSE}
&=
\sqrt{
\frac{1}{N}
\sum_{i=1}^{N}
\left(Y_i-\hat{Y}_i\right)^2
}.
\end{align}

Because forecasting errors during periods of concentrated screening activity are particularly consequential for checkpoint management, performance is also evaluated within high-throughput subsets. For \(q\in\{0.75,0.90\}\), let \(Q_q\) denote the corresponding quantile of observed throughput in the evaluation set, and define

\begin{equation}
\mathcal{I}_q =
\left\{
i:Y_i\geq Q_q
\right\}.
\end{equation}

Peak-period WMAPE is calculated as

\begin{equation}
\mathrm{WMAPE}_q
=
\frac{
\sum_{i\in\mathcal{I}_q}
\left|Y_i-\hat{Y}_i\right|
}{
\sum_{i\in\mathcal{I}_q}Y_i
}
\times 100\%,
\qquad
q\in\{0.75,0.90\}.
\label{eq:peak_wmape}
\end{equation}

Corresponding peak-period MAE values are also reported. For stochastic neural models, evaluation results are summarized using the mean and standard deviation across the five seeds.

\section{Results}
\label{sec:results}

\subsection{Direct 6 h Forecasting Performance}

\begin{figure}[t]
\centering
\includegraphics[width=0.92\linewidth]{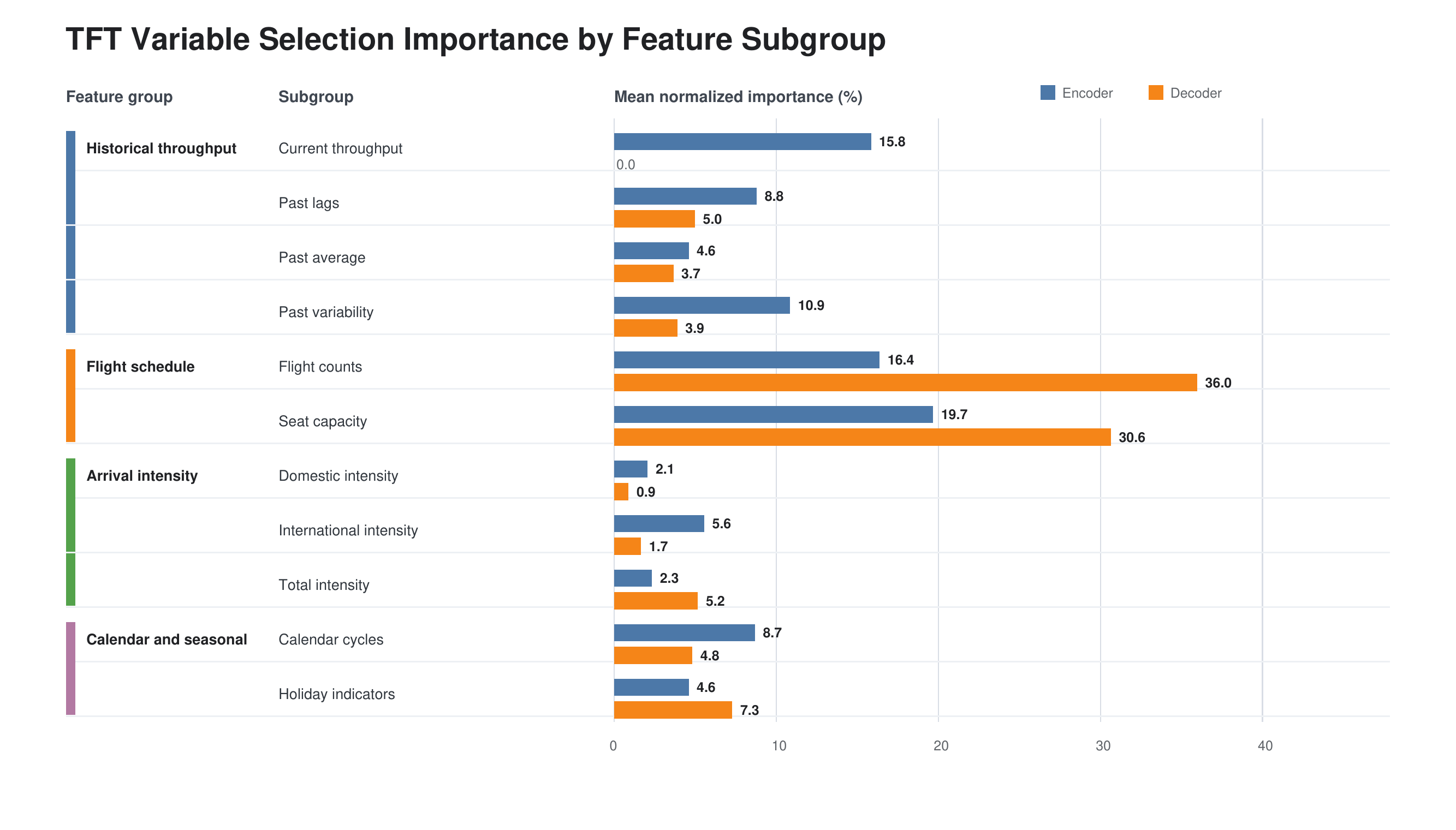}
\caption{TFT Variable Importance}
\label{fig:tft_attention_importance}
\end{figure}

Table~\ref{tab:baseline} reports the direct 6 h forecasting results. The proposed TFT model achieves the lowest overall error among the evaluated neural models, with an overall MAE of 317.87 and WMAPE of 9.33\%. Compared with RNN and LSTM, TFT reduces overall MAE by 19.14\% and 13.49\%, respectively. The improvement is also observed under high throughput conditions, where TFT obtains the lowest Peak75 MAE and Peak75 WMAPE. The corresponding encoder and decoder variable-selection importance scores are presented in Figure~\ref{fig:tft_attention_importance}.

\begin{table}[t]
\centering
\caption{Machine Learning baselines on ATL for 6 hour}
\label{tab:baseline}
\begin{tabular}{lrrrrrr}
\hline
Model & Overall & Overall & Peak75 & Peak75 & Peak90 & Peak90 \\
      & MAE & WMAPE & MAE & WMAPE & MAE & WMAPE \\
\hline
RNN  & 393.10 & 12.16\% & 498.18 & 9.17\% & 1013.05 & 17.41\% \\
LSTM & 367.45 & 11.37\% & 461.17 & 8.48\% & 866.69  & 14.89\% \\
TFT  & 317.87 & 9.33\%  & 437.39 & 7.22\% & 491.66  & 7.07\% \\
\hline
\end{tabular}
\end{table}

\subsection{Rolling Chunk Length Selection}

Table~\ref{tab:rolling} reports the 24 h rolling forecast results under different chunk lengths. The overall best performance is obtained when \(h=6\), with MAE of 362.62 and WMAPE of 10.54\%. In contrast, the one shot 24 h forecast has higher overall error, suggesting that recursive short horizon forecasting is more effective than predicting the full daily horizon at once. Based on the overall accuracy and operational update frequency, \(h=6\) is used as the representative rolling configuration for the long horizon analysis.

\begin{table}[t]
\centering
\caption{Rolling chunk length comparison for a 24 h planning horizon}
\label{tab:rolling}
\small
\begin{tabular}{rrrrrrr}
\hline
\(h\) & \multicolumn{2}{c}{Overall} & \multicolumn{2}{c}{Peak75} & \multicolumn{2}{c}{Peak90} \\
\cline{2-7}
 & MAE & WMAPE & MAE & WMAPE & MAE & WMAPE \\
\hline
1  & 370.00 & 10.76\% & 496.97 & 8.23\% & 537.80 & 7.94\% \\
3  & 365.54 & 10.63\% & 473.87 & 7.85\% & 514.49 & 7.59\% \\
6  & 362.62 & 10.54\% & 496.42 & 8.22\% & 516.08 & 7.62\% \\
8  & 375.50 & 10.92\% & 506.61 & 8.39\% & 552.59 & 8.16\% \\
12 & 400.42 & 11.64\% & 521.69 & 8.64\% & 555.51 & 8.20\% \\
24 & 388.95 & 11.31\% & 497.61 & 8.24\% & 568.81 & 8.40\% \\
\hline
\end{tabular}
\end{table}

\subsection{Long Horizon Rolling Forecasting}

Table~\ref{tab:long_horizon_results} evaluates the selected \(h=6\) rolling configuration over planning horizons from 24 to 96 h. Overall MAE remains
between 357.10 and 375.10 passengers, while WMAPE varies within \(10.60\%\)--\(11.04\%\). The highest overall error occurs at \(K=48\), after which both metrics decrease. Neither metric exhibits a monotonic relationship with \(K\), providing no evidence of systematic aggregate error amplification
as additional rolling blocks are introduced. However, because a complete future window is required, the \(24\), \(48\), \(72\), and \(96\) h evaluations
contain 25, 19, 13, and 8 valid origins per seed, respectively. The results should therefore be interpreted as horizon specific summaries rather than paired comparisons over identical airport hours.

Absolute errors are higher during periods of concentrated throughput. From \(K=24\) to \(K=96\), Peak75 MAE increases from 518.48 to 549.41 passengers,
while Peak90 MAE increases from 532.69 to 555.50 passengers. Nevertheless, Peak75 and Peak90 WMAPE remain within \(8.59\%\)--\(9.28\%\) and
\(7.86\%\)--\(8.48\%\), respectively. The lower peak period WMAPE values reflect the larger observed throughput in the denominator and do not indicate
smaller passenger count errors. Peak thresholds are calculated within each horizon specific evaluation sample.

\begin{figure}[t]
\centering
\includegraphics[width=0.92\linewidth]{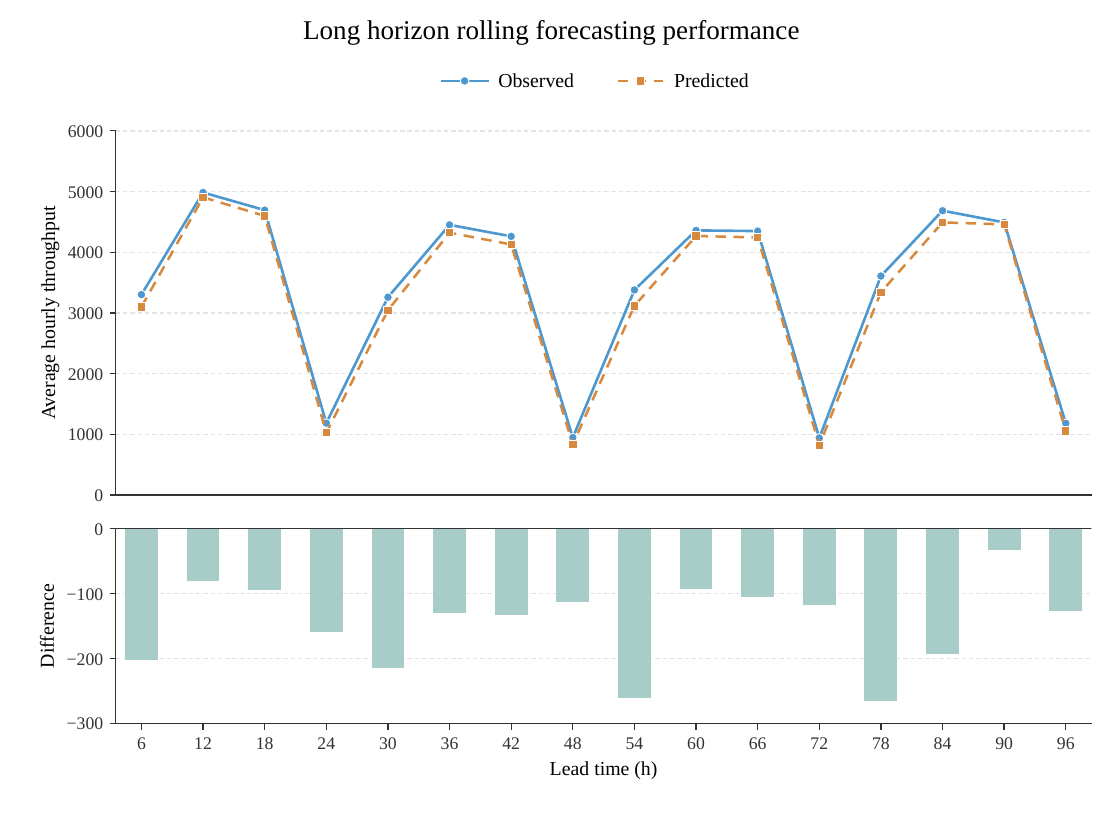}
\caption{Long horizon rolling forecasting performance with h = 6}
\label{fig:rolling_96h}
\end{figure}

Figure~\ref{fig:rolling_96h} complements these aggregate metrics by presenting the \(K=96\) forecast at successive 6 h lead intervals. The predicted trajectory reproduces the recurring daily peaks and troughs observed over the four day horizon, indicating that the dominant intraday structure is preserved during recursive forecasting. Predicted throughput remains below observed throughput during most intervals, revealing a modest underprediction tendency. However, the difference does not widen systematically with lead time, which is consistent with the nonmonotonic errors reported in Table~\ref{tab:long_horizon_results}.

\begin{table}[H]
\centering
\caption{Long horizon rolling forecasting performance with \(h=6\).}
\label{tab:long_horizon_results}
\small
\begin{tabular}{rrrrrrr}
\hline
\(K\) & \multicolumn{2}{c}{Overall} & \multicolumn{2}{c}{Peak75} & \multicolumn{2}{c}{Peak90} \\
\cline{2-7}
 & MAE & WMAPE & MAE & WMAPE & MAE & WMAPE \\
\hline
24 & 366.16 & 10.64\% & 518.48 & 8.59\% & 532.69 & 7.86\% \\
48 & 375.10 & 11.04\% & 539.71 & 9.04\% & 544.26 & 8.12\% \\
72 & 367.58 & 10.99\% & 543.59 & 9.25\% & 563.52 & 8.48\% \\
96 & 357.10 & 10.60\% & 549.41 & 9.28\% & 555.50 & 8.32\% \\
\hline
\end{tabular}
\end{table}

\section{Findings and Discussion}

The results indicate that the TFT model provides a more effective representation of TSA screening-load dynamics than the recurrent neural-network baselines. Rather than treating checkpoint throughput as a purely historical time-series pattern, the proposed framework combines observed throughput history with known future schedule information and schedule-derived arrival-intensity features. This structure is important because TSA screening demand is shaped by both recurring temporal patterns and future flight activity. The superior performance of TFT in the direct short-horizon experiment suggests that the model is better able to capture nonlinear interactions among historical throughput, calendar effects, flight schedules, and expected pre-departure arrival patterns. The finding therefore supports the central premise of the study: checkpoint throughput forecasting should be formulated as a schedule-informed operational forecasting problem, not simply as a generic passenger-volume prediction task.

The high-throughput evaluation further strengthens the operational relevance of the proposed approach. In checkpoint management, average forecasting accuracy is not sufficient because the consequences of error are uneven across the operating day. Errors during low-volume periods may have limited operational consequences, whereas errors during peak screening periods can affect lane-opening decisions, Transportation Security Officer deployment, queue formation, and passenger experience. The results show that TFT performs particularly well under high-throughput conditions, suggesting that the model is not merely fitting routine daily patterns but is also better able to anticipate periods of concentrated screening demand. This is especially important for airports such as ATL, where screening pressure can change rapidly within the day and where proactive planning depends on identifying when demand peaks are likely to occur.

The rolling forecast results show that short-horizon forecasts can be extended into practical planning windows through recursive updating. The comparison of rolling chunk lengths suggests that forecasting design involves a tradeoff between accuracy, responsiveness, and operational usability. Very short chunks allow frequent updates and may respond more quickly to recent demand changes, but they can be less convenient for planning workflows. Longer chunks are simpler to implement but may be less responsive to intraday variation. The selected rolling configuration provides a practical middle ground, supporting daily and multi-day screening-load planning without relying on a single long-horizon prediction. The long-horizon evaluation further suggests that recursive forecasting does not lead to substantial error amplification in this case study. This finding is meaningful because airport checkpoint planning often requires both near-term tactical forecasts and advance awareness of expected demand conditions over several days. However, the results should be interpreted as evidence of stable throughput forecasting, not as direct evidence of improved wait-time prediction or staffing efficiency.

Several limitations of the present study should be interpreted alongside corresponding directions for future research. First, the prediction target in this study is observed TSA throughput, which represents the number of passengers actually processed through screening during each hour. Although this is an appropriate operational measure of realized checkpoint load, it may not fully capture the underlying passenger arrival process. Observed throughput can be affected not only by passenger demand, but also by checkpoint capacity, lane availability, staffing levels, screening technology, operating procedures, and temporary disruptions. As a result, the model forecasts realized screening activity rather than unconstrained passenger arrivals or queue formation. Future research should integrate throughput data with additional operational indicators, such as lane openings, staffing levels, service rates, queue lengths, or wait-time observations. Such integration would allow researchers to separate passenger arrival demand from screening capacity constraints and better evaluate how forecasted demand translates into congestion, staffing needs, and passenger service reliability. 

In addition, due to data availability, this study models TSA checkpoint throughput at the hourly level. Although hourly forecasts are useful for broad staffing and lane-planning decisions, checkpoint congestion can form over much shorter intervals. Future research should examine whether higher-temporal-resolution data, such as 10-minute or 15-minute checkpoint throughput, can improve the model’s ability to capture short-lived demand surges, peak buildup, and queue-formation dynamics. Finer-resolution data would also allow the schedule-derived arrival-intensity features to be calibrated more precisely to passenger show-up behavior before departure. However, higher-resolution forecasting may introduce additional noise and data-sparsity challenges, especially during low-volume periods or at smaller airports. Therefore, future work should evaluate not only whether finer temporal resolution improves predictive accuracy, but also whether it provides meaningful operational value for checkpoint staffing, lane-opening decisions, and real-time congestion management. 

Furthermore, the schedule-derived arrival-intensity features are based on fixed lead-time assumptions for domestic and international flights. This representation is useful because it translates scheduled departures into expected pre-departure screening-load patterns, but passenger show-up behavior is likely to vary across airports, airlines, routes, passenger types, time of day, day of week, baggage status, and perceived uncertainty about processing time. A fixed lead-time kernel may therefore underrepresent important behavioral heterogeneity in when passengers arrive at the checkpoint. Future research should estimate airport-specific, route-specific, or time-varying show-up profiles using richer operational data. For example, passenger arrival profiles could be calibrated using checkpoint scan data at finer temporal resolution, airline check-in records, boarding data, mobile-device-based terminal movement data, or observed wait-time patterns. This would allow schedule-derived features to better reflect actual passenger behavior and improve the model’s ability to assign future screening demand to the correct operating intervals.

\section{Conclusion}

This study proposed a schedule-informed framework for hourly TSA checkpoint throughput prediction. By translating scheduled departures into pre-departure arrival-intensity features and incorporating them into a Temporal Fusion Transformer model, the framework addresses a central operational challenge in checkpoint forecasting: screening demand is generated before flight departure and must be predicted at the time scale relevant to staffing and lane-management decisions. The empirical results demonstrate the value of combining historical throughput patterns with known future schedule information for short-horizon and rolling multi-horizon forecasting. Beyond improving predictive performance, the main contribution of the study is to provide an operationally meaningful way to connect flight schedules with checkpoint screening-load prediction. This is important because effective checkpoint planning depends not only on estimating passenger volume, but also on anticipating when screening pressure is likely to occur.

To conclude, this transformer-based framework should be viewed as a forecasting layer that can support, but not replace, broader checkpoint management systems. In practice, throughput forecasts can be integrated with queueing models, staffing rules, lane-capacity assumptions, and simulation or optimization tools to evaluate service reliability and resource-allocation strategies. Future research should extend the framework using finer-resolution data, richer operational measures, and airport-specific passenger show-up profiles to improve its behavioral realism and practical applicability across different airport contexts.

\section{Acknowledgments}

This work used the Anvil supercomputer at Purdue University through the ENG260007 allocation from the Advanced Cyberinfrastructure Coordination Ecosystem: Services \& Support (ACCESS) program, which is supported by the U.S. National Science Foundation under Grants 2138259, 2138286, 2138307, 2137603, and 2138296 \citep{song2022anvil,boerner2023access}.

ChatGPT-5 was used solely to assist with grammatical and formatting checks. The authors reviewed and verified all AI-assisted content and take full responsibility for the final manuscript.

\section{Author Contributions}

The authors confirm contribution to the paper as follows: study conception and design: Yinxiao Zhang, Dr. Sen Wang; data collection: Yinxiao Zhang, Dr. Sen Wang, Dr. Yi Gao; analysis and interpretation of results: Yinxiao Zhang, Dr. Sen Wang; draft manuscript preparation: Yinxiao Zhang, Dr. Sen Wang, Dr. Yi Gao. All authors reviewed the results and approved the final version of the manuscript.

\section{Declaration of Conflicting Interests}

The authors declared no potential conflicts of interest with respect to the research, authorship, and/or publication of this article.

\section{Funding}

The authors disclosed no financial support for the research, authorship, and/or publication of this article.

\newpage
\printbibliography[title={REFERENCES}]
\end{document}